\documentclass[10pt,twocolumn,letterpaper]{article}

\usepackage{iccv}
\usepackage{times}
\usepackage{epsfig}
\usepackage{graphicx}
\usepackage{amsmath}
\usepackage{amssymb}
\usepackage{multirow}
\usepackage{caption}
\usepackage{subcaption}
\usepackage{threeparttable}
\usepackage{latexsym}
\usepackage{booktabs}

\usepackage[breaklinks=true,bookmarks=false]{hyperref}

\iccvfinalcopy 

\def\iccvPaperID{4} 

\ificcvfinal\fi

\newcommand{\ourmodel}{LOGOS}
\newcommand*\samethanks[1][\value{footnote}]{\footnotemark[#1]}

\begin{document}



\title{Localize, Group, and Select: Boosting Text-VQA by Scene Text Modeling}



\author{Xiaopeng Lu\thanks{Equally contributed}, Zhen Fan\samethanks, Yansen Wang\samethanks, Jean Oh, Carolyn P. Rosé \\
  Language Technologies Institute, Carnegie Mellon University, \\
  5000 Forbes Avenue, Pittsburgh PA, 15213 \\
  \texttt{\{xiaopen2, zhenfan, yansenwa, jeanoh, cp3a\}@andrew.cmu.edu} 
  \\
  }

\maketitle

\begin{abstract}

As an important task in multimodal context understanding, Text-VQA (Visual Question Answering) aims at question answering through reading text information in images. It differentiates from the original VQA task as Text-VQA requires large amounts of scene-text relationship understanding, in addition to the cross-modal grounding capability. In this paper, we propose \textbf{Lo}calize, \textbf{G}r\textbf{o}up, and \textbf{S}elect (LOGOS), a novel model which attempts to tackle this problem from multiple aspects. LOGOS leverages two grounding tasks to better \textit{localize} the key information of the image, utilizes scene text clustering to \textit{group} individual OCR tokens, and learns to \textit{select} the best answer from different sources of OCR (Optical Character Recognition) texts. Experiments show that LOGOS outperforms previous state-of-the-art methods on two Text-VQA benchmarks without using additional OCR annotation data. Ablation studies and analysis demonstrate the capability of LOGOS to bridge different modalities and better understand scene text.
\end{abstract}

\section{Introduction}
\label{sec:intro}

Text-VQA, which calls for a model to answer questions based on scene text in Visual Question Answering (VQA) settings, has attracted much attention in recent years. The task is set to evaluate a model's capability to detect and understand textual information from images, and to make inference between cross-modal facts. 
Various high-quality Text-VQA benchmarks have been released in recent years \cite{singh2019towards, biten2019scene, mishra2019ocr, wang2020general}, as well as multiple proposed methods to address this task.

\begin{figure}[!t]
    \centering
    \includegraphics[width=\linewidth]{figure/books.jpg}
    \caption{Example of Text-VQA. The question is ``What is the title of the book in the middle?".}
   \label{fig:example}
\end{figure}

Previous works have followed the pipeline of first extracting the embedded scene texts, and then jointly generating the answer based on input question, extracted texts, and visual contents \cite{singh2019towards, kant2020spatially, yang2020tap}. However, simply selecting tokens from detected scene text is often not enough.
Text-VQA differs from other multi-modal language analysis tasks (multi-modal sentiment analysis, image retrieval, etc.) where there is a direct connection between task and textual information, and the text itself is often complete, reliable, and can be emphasized over other modalities \cite{wang2019words,rahman2020integrating}. In the Text-VQA scenario, things become much more complicated because: (1) The scene text itself may not contain enough information to fulfill the information need, especially when the given question focuses on certain visual characteristics (e.g. \textit{name written in blue}) or certain objects (e.g. \textit{text written on the boat}). (2) The layout of scene text is noisy in real-life images: a single sentence may be split into different rows in the image, which makes it difficult for the current OCR system to recognize them as a full sentence or paragraph. (3) OCR systems may make detection errors, especially on real-life images from very diverse domains where different OCR systems with different tendencies may excel or struggle, which is a natural bottleneck for this task. 

Figure~\ref{fig:example} shows a typical example of Text-VQA. In order to answer the question ``\textit{What is the title of the book in the middle?}" correctly, the model needs to overcome several obstacles. It should bridge the question with corresponding textual or visual characteristics (\textit{the book in the middle}), properly group the ``lines" of words on the book covers to form book and author names, and avoid possible OCR detection errors and choose the most reasonable and reliable OCR output from multiple sources. 

Considering these difficulties, in this paper, we propose a model named LOGOS (Localize, Group, and Select) for better scene text understanding. Concretely, we \textbf{localize} the question to its Region of Interest (ROI) by (1) introducing a visual grounding task to connect question text and image regions and (2) connecting regions with text modalities using object tokens as explicit alignments. Also, we \textbf{group} the individual text pieces via unsupervised, position-based clustering, and provide such positional information to the model as layout representation. Finally, we utilize OCR systems of the different capability to model words in different granularity, and train LOGOS to dynamically \textbf{select} the answer from multiple noisy OCR sources. Experiments on two benchmarks confirm that our model benefits from better modeling the text modality and can outperform the state-of-the-art models without additional OCR data annotation. 

The main contributions of this paper are the following:

\begin{itemize}
\setlength{\itemsep}{1.5pt}
\setlength{\parsep}{1.5pt}
\setlength{\parskip}{1.5pt}
    \item A novel Text-VQA model that effectively grounds different modalities and denoise low-quality OCR inputs
    \item State-of-the-art results over two benchmark Text-VQA datasets 
    \item Detailed analysis on the advantages of LOGOS in cross-modal grounding and scene-text understanding
\end{itemize}

\section{Related Works}
\label{sec:related}

The Text-VQA task aims at reading and understanding the text captured in the images to answer visual questions. Although the importance of involving scene texts in visual question answering tasks was originally emphasized by \cite{bigham2010vizwiz}, due to the lack of available large-scale datasets, early development of question answering tasks related to embedded text understanding was limited in narrow domains such as bar-charts \cite{kafle2018dvqa} or diagrams \cite{kembhavi2016diagram, kembhavi2017you}. The first large-scale open-domain dataset of the Text-VQA task, TextVQA, was introduced by \cite{singh2019towards}, followed by several similar works including STVQA \cite{biten2019scene}, OCR-VQA \cite{mishra2019ocr} and EST-VQA \cite{wang2020general}. 

\begin{figure*}[!th]
    \centering
    \includegraphics[width=\linewidth]{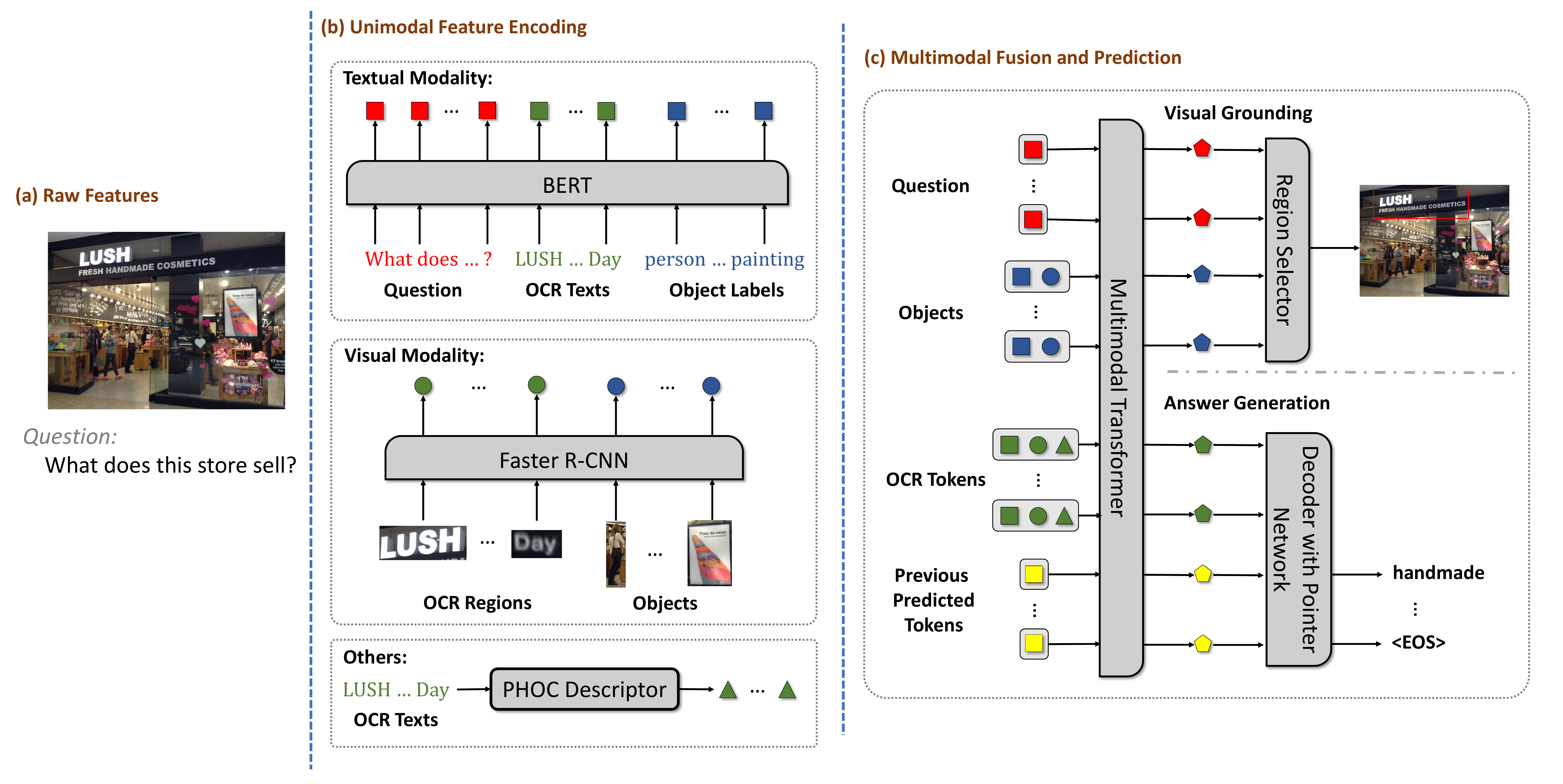}
    \caption{Overview of the LOGOS model. It illustrates how the raw features from the image and the question are processed step by step. Marks of the same \textbf{shape} represent features of the same modality, and marks of the same \textbf{color} represent features from the same text token or image region. }
    \label{fig:model}
\end{figure*}
Recent studies \cite{singh2019towards,hu2020iterative,kant2020spatially,yang2020tap,gao2020structured,gao2020multi,liu2020cascade,han2020finding} have proposed several models and network architectures for the Text-VQA task. Introduced together with the TextVQA dataset, the LoRRA \cite{singh2019towards} model is built upon Pythia \cite{jiang2018pythia} with an OCR module added to detect and recognize the scene texts. M4C \cite{hu2020iterative} first models Text-VQA as a multimodal task and uses a multimodal transformer to fuse different features over a joint embedding space. Also, it attaches a pointer network that can dynamically copy words from OCR systems. SA-M4C \cite{kant2020spatially} added spatial information between objects, OCR, and question tokens to implicitly capture their relationships based on M4C to get further improvement. TAP \cite{yang2020tap} proposed to pretrain the model on several auxiliary tasks such as masked language modeling (MLM) and relative position prediction (RPP). It also leverages additional large-scale OCR datasets to enhance its ability to capture the contextualized information of scene text. 

Although some previous works reported better results obtained by purely changing the OCR system \cite{kant2020spatially, yang2020tap}, they either don't fully realize the potential of modeling the text modality with information from the visual modality or use expensive large-scale OCR data for pretraining. In this paper, we look deeper into other solutions to ground and refine features in the textual modality within existing datasets to facilitate the understanding of scene text in the multimodal fusion process.
\section{LOGOS Model}
\label{sec:method}


Figure \ref{fig:model} demonstrates the \ourmodel~model structure, upon which we attempt to bridge the modalities of the question, image, and OCR text with different approaches. The focus of our model is three-fold:
\begin{itemize}
\setlength{\itemsep}{1.5pt}
\setlength{\parsep}{1.5pt}
\setlength{\parskip}{1.5pt}
    \item Localize ROI by question-visual pretraining and question-OCR text modeling
    \item Group related scene text tokens by clustering OCR layout information
    \item Select OCR tokens dynamically from multiple noisy OCR systems
\end{itemize}

\subsection{Model Architecture}
Considering Text-VQA as a typical multi-modal task with inputs from several different modalities, our model utilizes the hybrid fusion technique, where the first step is to generate separate unimodal representations of different modalities. We use BERT \cite{devlin2018bert} and Faster R-CNN \cite{ren2015faster} to encode the text and image respectively. Following previous works \cite{hu2020iterative, yang2020tap}, we also include the pyramidal histogram of characters (PHOC) representations~\cite{almazan2014word} for the OCR tokens. 

To let the model aware of ROI given a certain question, we leverage question-visual pretraining before Text-VQA, and question-OCR modeling during Text-VQA training, with object tokens to help align two modalities, with details shown in Section~\ref{sec:roi}.

With the aforementioned inputs, we add grouping information for OCR tokens based on their locations. The details of the clustering algorithm are introduced in Section~\ref{sec:ocr-cluster}.

As shown in Figure~\ref{fig:model}(b), after uni-modal features are obtained, the representations of the same OCR and object tokens from different modalities are concatenated respectively. We feed the textual representation of the question and the fused representation of detected objects and OCR tokens into a multi-modal transformer. We utilize an answer decoder with a pointer network for output generation and design a selection framework for multiple OCR sources, which we discuss in Section~\ref{sec:selector}. 


\subsection{ROI Localization}
\label{sec:roi}
To successfully predict the answer given the question information, a robust model should be capable of first pointing out the specific region that is most related to the question, then relying on the OCR text of that \textit{specific} region to generate the output. 

It is the most desirable case if we have large amounts of alignment data between question and OCR regions to learn the question-OCR grounding. Since there is no publicly available large-scale question-OCR dataset, we view the question-OCR grounding problem as a two-stage learning problem. Specifically, we first learn the \textit{general} grounding between text and image regions usin existing region-description datasets, for example, Visual Genome~\cite{krishna2016visual}. Second,  following the intuition from Oscar~\cite{li2020oscar}, we use the object tokens as textual ``guidance" to help the model learn to ground between question and OCR regions.



\subsubsection{Question-Visual Pretraining}
\label{sec:visual-grounding}

To better equip the model with this ability, we select referral expression selection as a pretraining task for question-visual grounding purposes. Concretely, given an image with a description and $n$ non-overlapping bounding boxes, the model learns to predict which bounding box is the one that is best aligned with the description. As shown in Figure~\ref{fig:model}, we reuse most parts of the network but add an extra classification layer for candidate prediction. As shown in Figure~\ref{fig:model} (c), similar to \ourmodel~ encoding structure, we use question encodings and object encodings as input to the multi-modal transformer module. We train the question-visual pretraining task using cross-entropy loss. 


\subsubsection{Question-OCR Modeling}
\label{sec:text-grounding}
Question-Visual Pretraining learns general grounding between question and object regions. In the second stage, we bring object tokens together with object region representations to bridge the gap between question tokens and unseen OCR regions. As shown in Figure~\ref{fig:model} (b), we concatenate object label tokens together with question and OCR texts and feed them into a uni-modal BERT encoder. Uni-modal BERT encoder models contextualized representation jointly among question, objects, and OCR texts. After passing through the uni-modal BERT layer, we follow the first stage pretraining routine with OCR concatenated representation as an additional input source to the multi-modal transformer encoder. The injection of object information in ~\textit{both} uni-modal and multi-modal transformers can be seen as a bridge to help learn the grounding information between question and OCR inputs, in the case where large amounts of question-OCR data are not available. 

\begin{figure}[ht!]
\begin{subfigure}[b]{0.23\textwidth}
    \includegraphics[width=\textwidth]{figure/clus_before.png}
    \caption{a. OCR detection}
    \label{fig:clus_before}
\end{subfigure}
\begin{subfigure}[b]{0.23\textwidth}
    \includegraphics[width=\textwidth]{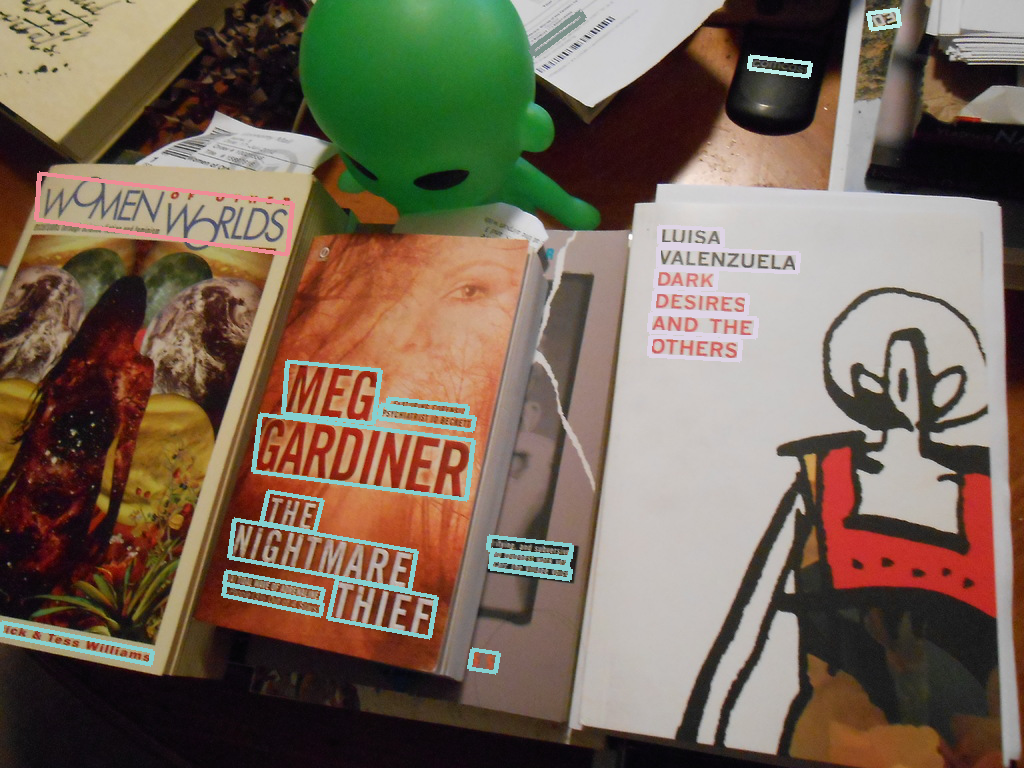}
    \caption{b. Clustering results}
    \label{fig:clus_after}
\end{subfigure}
\caption{Example of position-based cluster. Each box represents a line detected by OCR. Same box color in Figure~\ref{fig:clus_after} represents same clusters.}
\end{figure}

\subsection{Scene Text Clustering and Modeling}
\label{sec:ocr-cluster}
Context understanding has always been important in language processing and question answering. This problem becomes even more vital in Text-VQA, where the evidence of clear separation of text groups lies in the visual modality rather than text. The detection scope of current OCR systems also mainly remains at individual token level or ``line" level (grouping tokens aligned closely forming a line). 

Figure~\ref{fig:clus_before} shows an example of raw detection results by an OCR system. For most text groups such as names and authors of the books, its words are split into multiple lines and thus detected individually. Simply aligning text pieces by their generated sequence, and not providing extra visual evidence will often lead to misordered text and incomplete/incorrect context modeling. 

A straightforward idea would be to extract potential text groups based on visual modality evidence, which may serve as additional spatial information along with raw OCR bounding box coordinates, and provide weak evidence for token realignment or token context to form complete sentences in the text domain. 

In LOGOS, we perform unsupervised clustering of the OCR text bounding boxes during data preparation. Given an image containing $m$ lines $\{L_1,\dots,L_m\}$ detected by OCR, with $L_i=t_{i1}t_{i2}\dots t_{il_i}$ containing $l_i$ tokens, we define the distance between two bounding boxes as the minimum distance between any two points on the two bounding boxes, and perform clustering using DBSCAN~\cite{hahsler2019dbscan} on the bounding boxes of all lines $\{ B_i  \}$. Through clustering, each token $t_{ij}$ is assigned extra spatial hierarchical information in its cluster $C_i$, line $i$ and token $j$. We generate a $d$ dimensional sinusoidal embedding as a positional representation for each attribute to form the overall descriptor $v^{sp}_{ij}$, and concatenate it as part of the final OCR representation. 
\begin{align}
    C_i &= \text{DBSCAN}(\{B_1\dots B_m\},\epsilon)[i] \\
    v^{sp}_{ij} &= \text{Pos}({C_i,d})\| \text{Pos}({i,d})\| \text{Pos}({j,d})
\end{align}
Where $\text{Pos}(i,d)$ is the $d$-dim sinusoidal positional embedding at position $i$. Figure~\ref{fig:clus_after} shows a sample result of this clustering approach.

\subsection{OCR Source Selection}
\label{sec:selector}
As aforementioned, current OCR systems are not robust enough to perfectly extract high-quality scene text from images. Different systems also result in different kinds of errors: while a careless OCR system will not be capable of finding all text in the images correctly and make more detection error, a meticulous OCR system may detect too much text from details inside the image, which will increase the difficulty of localizing the question to the correct region. 
This opens up the possibility of minimizing OCR error and maximizing reasoning effectiveness by combining different OCR sources.

We modify the training and predicting process order to best utilize different OCR systems. Given OCR scene text from $k$ systems, during the decoding stage, $k$ independent answers are generated separately. LOGOS then calculates the confidence score for the $i$-th answer $S^i=t_1^it_2^i...t_l^i$ as
\begin{align}
    Sc&ore(S^i) = \mathcal{P}(S^i) \nonumber\\
    &= \mathcal{P}(t_1^i|x)\prod_{j=2}^{l}\mathcal{P}(t_j^i|x, t_1^i, ..., t_{j-1}^i),
\end{align}
where $i\in[k]$ and $x$ is the input of other features including visual features and questions. The answer with the highest score is selected as the final answer. During training, all $k$ OCR sources are trained equally, which also serves as extra training data for the learning process.


\section{Experiments}
\label{sec:expr}
\subsection{Datasets and Evaluation Metrics}
In this paper, we use TextVQA~\cite{singh2019towards} and STVQA~\cite{biten2019scene}, two commonly used scene-text VQA datasets as our main testbeds. Visual Genome dataset~\cite{krishna2016visual} is used for the referral expression auxiliary training. 

\paragraph{TextVQA} is the first large-scale Text-VQA dataset with 28,408 images sampled from the Open Image Dataset~\cite{krasin2017openimages}. A total of 45,336 questions related to the text information in the image were answered by annotators. For each question-image pair, 10 answers are provided by different annotators. The accuracy normalized by weighted voting over the 10 answers is reported on this dataset. Following previous settings~\cite{singh2019towards,hu2020iterative}, we split the dataset into 21,953, 3,166, and 3,289 images respectively for train, validation, and test set.

\paragraph{STVQA} is similar to the TextVQA dataset and it contains 23,038 images with 31,791 questions. We follow the setting from M4C~\cite{hu2020iterative} and split the dataset into train, validation, and test splits with 17,028, 1,893, and 2,971 images respectively. Compared with TextVQA dataset, the data source of STVQA is more diverse which includes data from Coco-text~\cite{veit2016coco}, Visual Genome~\cite{krishna2016visual}, VizWiz~\cite{gurari2018vizwiz}, ICDAR~\cite{karatzas2015icdar}, ImageNet~\cite{deng2009imagenet}, and IIIT-STR~\cite{mishra2013image} data. We report 2 metrics, accuracy and Average Normalized Levenshtein Similarity(ANLS) on this dataset.

\paragraph{Visual Genome} is a multi-modal grounding dataset with 108,077 images~\cite{krishna2016visual}. We use this dataset for the purpose of visual grounding joint training.  Specifically, region descriptions and their corresponding bounding boxes are used for generating multiple-choice data. In total, we generate 1,216,806 training pairs from the Visual Genome dataset.

 \begin{table*}[t]
 \centering
 \begin{threeparttable}
 \scalebox{1.0}{
 \begin{tabular}{c|l|c|c|c|c|c}
 \toprule
  \multicolumn{1}{c|}{\multirow{2}*{\textbf{OCR Data}}} &
  \multicolumn{1}{c|}{\multirow{2}*{\textbf{Method}}} & \multicolumn{2}{c|}{\textbf{TextVQA}} & \multicolumn{3}{c}{\textbf{ST-VQA}}   \\
  \cline{3-7}
  & & Acc.(val) & Acc.(test) & Acc.(val) & ANLS(val) & ANLS(test) \\
  \midrule
  External data$^\dagger$ &TAP~\cite{yang2020tap} & 54.71 & 53.97 & 50.83 & 0.598 & 0.597  \\
  \midrule
  \multirow{9}*{Original dataset only} & LoRRA~\cite{singh2019towards} & 26.56 & 27.63 & - & - & - \\
  &SAN+STR~\cite{biten2019scene} & - & - & - & - & 0.135 \\
  &M4C~\cite{hu2020iterative} & 40.55 & 40.46 & 38.05 & 0.472 & 0.462 \\
  &M4C+Azure OCR & 45.22 & - & 42.28 & 0.517 & 0.517 \\
  &SMA~\cite{gao2020structured} & 40.05 & 40.66 & - & - & 0.466 \\
  &CRN~\cite{liu2020cascade} & 40.39 & 40.96 & - & - & 0.483 \\
  &LaAP-Net~\cite{han2020finding} & 40.68 & 40.54 & 39.74 & 0.497 & 0.485 \\
  &SA-M4C~\cite{kant2020spatially} & 45.40 & 44.60 & 42.23 & 0.512 & 0.504 \\
  &TAP~\cite{yang2020tap}  & 49.91 & 49.71 & \textbf{45.29} & \textbf{0.551} & \textbf{0.543} \\
  &\textbf{\ourmodel~(Ours)}  & \textbf{50.79} & \textbf{50.65} & 44.10 & 0.535 & 0.522 \\
  \midrule
  \multirow{4}*{TextVQA+STVQA}  
  &LaAP-Net~\cite{han2020finding} & 41.02 & 41.41 & - & - & - \\
  &SNE~\cite{zhu2020simple} & 45.53 & 45.66 & - & - & 0.550 \\
  &TAP~\cite{yang2020tap}  & 50.57 & 50.71 & - & - & - \\
  &\textbf{\ourmodel~(Ours)} & \textbf{51.53} & \textbf{51.08} & \textbf{48.63} & \textbf{0.581} & \textbf{0.579} \\
  \bottomrule
 \end{tabular}}
 \begin{tablenotes}
 \footnotesize
 \item[$^\dagger$] Uses extra OCR annotations on the OCR-CC dataset, currently not publicly available, with TextVQA and STVQA. 
 \end{tablenotes}
 \end{threeparttable}
 \caption{Model performance on TextVQA and STVQA.}
 \label{tab:experiment_result}
 \end{table*}

\subsection{Experiment settings and Training Details}
Our model is implemented based on the framework of M4C~\cite{hu2020iterative}. For the visual modality, we follow the settings of M4C and use Faster R-CNN~\cite{ren2015faster} fc7 features of 100 top-scoring objects in the image detected by a Faster R-CNN detector pretrained on the Visual Genome dataset. The fc7 weights are fine-tuned during training. For the text modality, we use two OCR systems, Rosetta OCR~\cite{borisyuk2018rosetta} and Azure OCR
~\footnote{\url{https://azure.microsoft.com/en-us/services/cognitive-services/computer-vision}},
to recognize scene text. We use a trainable 3-layer BERT-base encoder~\cite{devlin2018bert} for text representation. We follow M4C and include pyramidal histogram of characters (PHOC) representations~\cite{almazan2014word} of OCR text. We perform scene text clustering on normalized bounding boxes using DBSCAN with $\epsilon=0.02$.


The candidate vocabulary for decoding consists of the top 5,000 frequent words from the answers in the training set, as well as the detected OCR tokens in the current image. The answer is generated with a pointer network~\cite{vinyals2015pointer}.

LOGOS contains 116M trainable parameters. During pretraining steps, we first pretrain LOGOS using the question-visual grounding task, then continue to train on the Text-VQA task. We set the batch size to 48 and train for 24,000 iterations for both pretraining and fine-tuning stages. The initial learning rate is set to 1e-4 with a warm-up period of 1000 iterations, and the learning rate decays to 0.1x after 14,000 and 19,000 iterations respectively. We use the checkpoints that achieved the best performance on the validation set for evaluation.

\subsection{Experiment Results}

Table~\ref{tab:experiment_result} lists the performance of \ourmodel~on the TextVQA dataset comparing to other baselines. 
We find that \ourmodel~outperforms all current baselines which do not use extra OCR data. 

We have several interesting observations from this table: (1) We see a huge gap between models using another OCR system (M4C+Azure OCR, SA-M4C, TAP, LOGOS) and models using the Rosetta system. The gap illustrates that previous performances on the Text-VQA task are severely limited by the quality of the scene texts detected. This proves the importance of better modeling and refining for the textual modality. (2) LOGOS shares the same idea of introducing auxiliary tasks for pretraining or joint training with TAP, and the results improve similarly although the training tasks are very different. However, LOGOS still outperforms TAP by 0.93\% on the TextVQA dataset because of better modeling of text features. (3) It's noteworthy that the current highest score from TAP is obtained by pretraining on other large-scale OCR datasets (OCR-CC) which are designed to better utilize scene text features in multimodal tasks. This idea is similar to ours and our model is fully compatible with pretraining on this dataset. We expect a further improvement of our model when it can get access to more OCR data. 

Looking at STVQA results, we can see that training using only STVQA data does not outperform the TAP model~\cite{yang2020tap}. This may due to the fact that STVQA contains more short-length answers, compared with TextVQA where the questions are relatively longer and harder to answer. As the dataset size of STVQA is small and the spatial relationship of this dataset is relatively simple, we observe that LOGOS suffers from overfitting in this dataset. As introducing the joint training with TextVQA dataset, we do see a huge improvement on the STVQA dataset: our model outperforms SNE, previously best STVQA model jointly trained on TextVQA dataset by 2.9 points in ANLS. This reveals that LOGOS can also achieve great performance in STVQA dataset with the additional training data from other Text-VQA datasets. 
\section{Analysis}
\label{sec:analysis}

\begin{table*}[t]
\centering
\label{tab:ablation}
\begin{tabular}{l|c|c|c|c|c}
\toprule
 \multirow{2}*{\textbf{\#}} & \textbf{Question-OCR } & \textbf{ Question-Visual} & \textbf{Scene Texts} &  \multirow{2}*{\textbf{Source(s)}} & \textbf{TextVQA} \\
 & \textbf{Modeling} & \textbf{Pretraining} & \textbf{Clustering} &  & \textbf{Acc.(val)}\\
 \midrule
  1 &   &  &  & A & 47.76 \\
  2 &   & \checkmark &  & A &48.26 \\
  3 &  \checkmark &  & & A &48.93 \\
  4 &  \checkmark &  & & R &40.84\\
  5 &    \checkmark &  & & A+R &49.79 \\
  6 &  \checkmark  & & \checkmark & A+R &50.40 \\ 
  7 &  \checkmark & \checkmark & \checkmark & A+R &\textbf{50.79} \\ 

\bottomrule 
\end{tabular}
\caption{Ablation studies of LOGOS on TextVQA dataset; A refers to Azure OCR and R refers to Rosetta OCR.}
\label{tab:ablation}
\end{table*}

\begin{figure*}[thp]
    \begin{subfigure}[b]{0.50\linewidth}
    \includegraphics[width=0.48\linewidth]{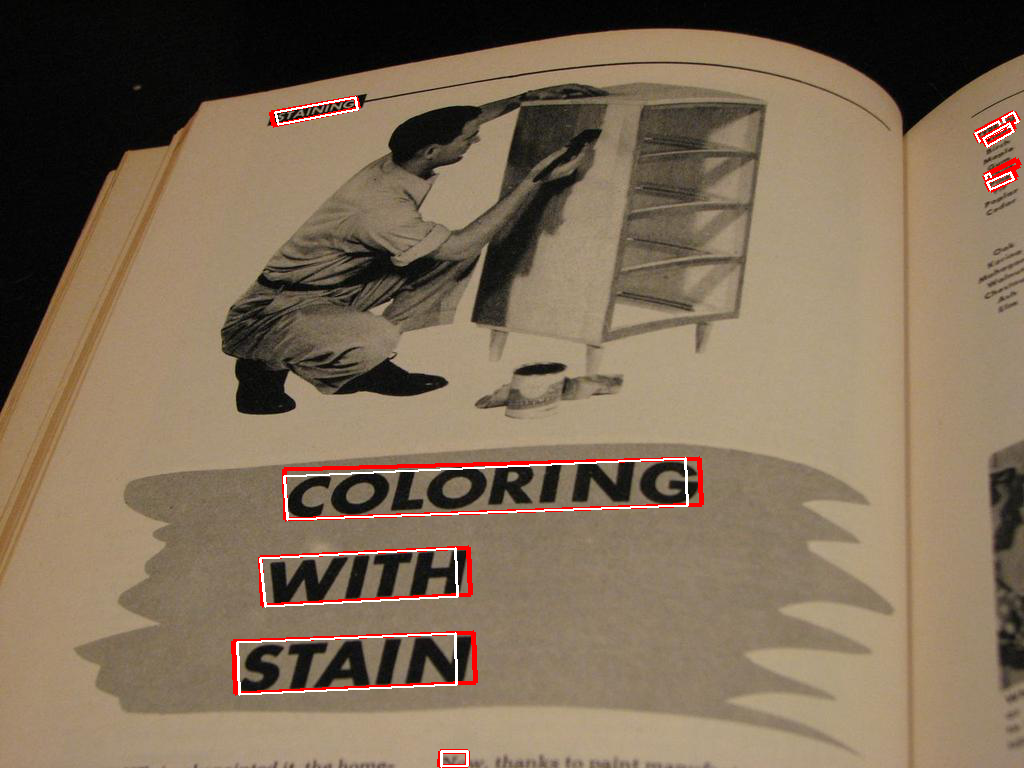}
    \includegraphics[width=0.48\linewidth]{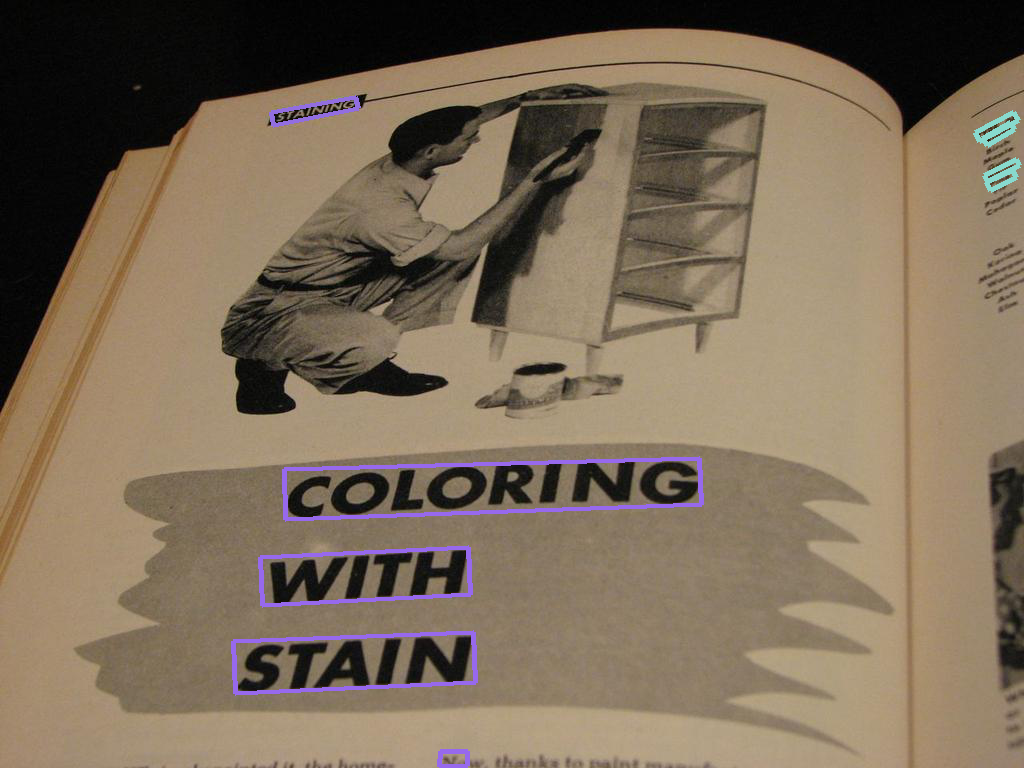}
    {
    \small
    \textbf{Question:} What are they coloring with this instructional book? \\
    \textbf{LOGOS (w/o Scene Text Clustering):} \textcolor{red}{coloring} \\
    \textbf{LOGOS (w. Scene Text Clustering):} \textcolor{blue}{stain}}
    \end{subfigure}
    \hspace{0.03\linewidth}
    \begin{subfigure}[b]{0.50\linewidth}
    \includegraphics[width=0.48\linewidth]{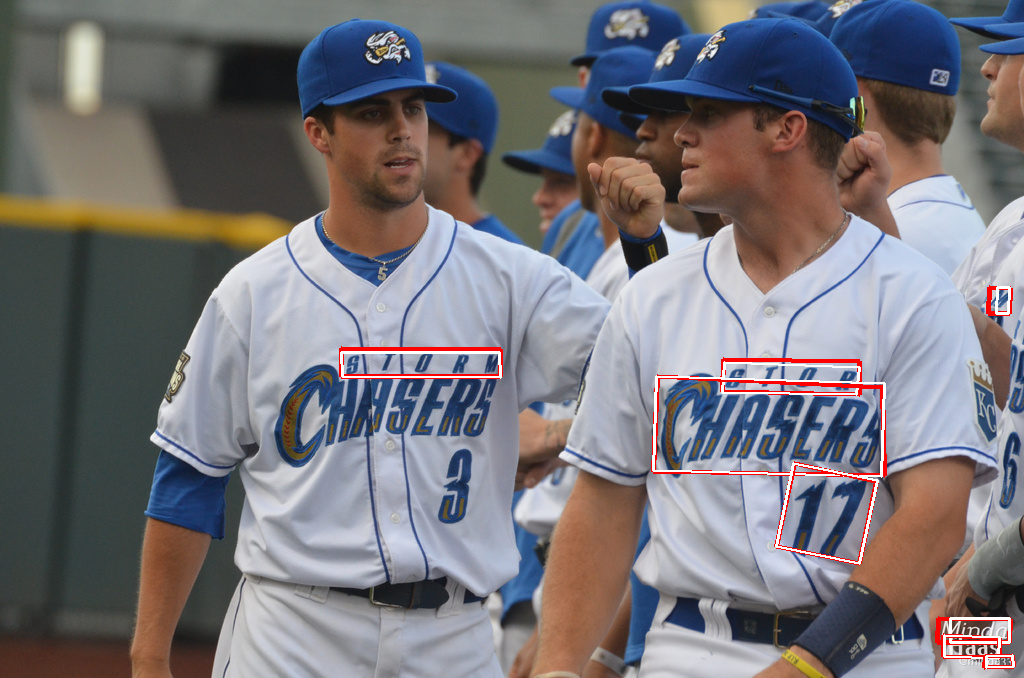}
    \includegraphics[width=0.48\linewidth]{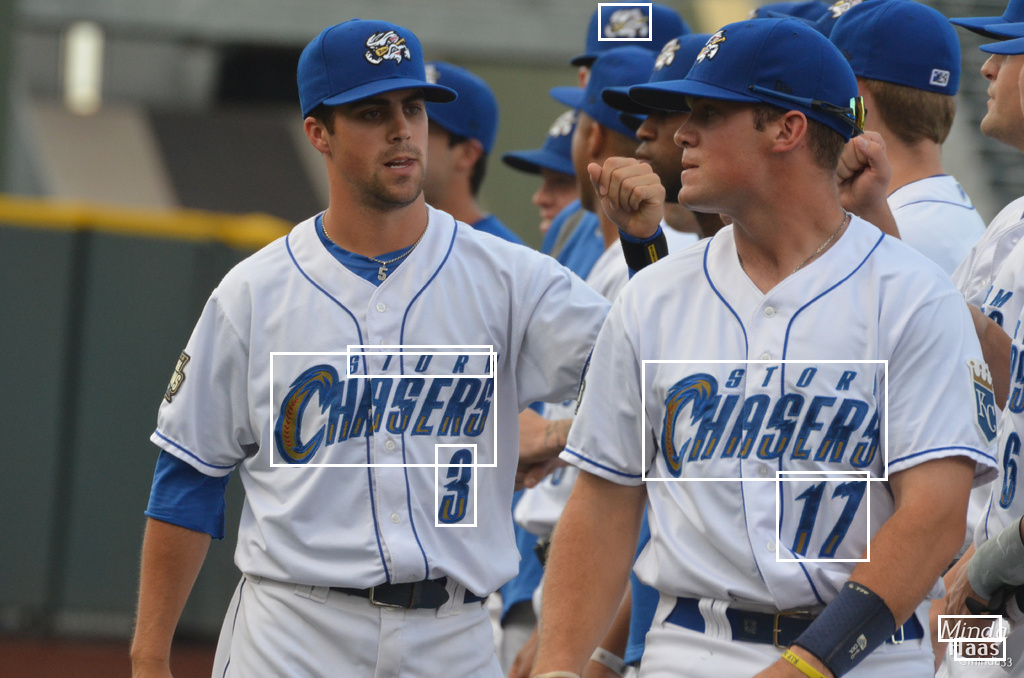}
    {
    \small
    \textbf{Question:} What is the number of the jersey on the left? \\
    \textbf{LOGOS (Azure):} \textcolor{red}{17} \\
    \textbf{LOGOS (Source Selection - Azure, Rosetta):} \textcolor{blue}{3} 
    }
    \end{subfigure}
  \caption{Qualitative comparisons between LOGOS and its ablated model predictions. \textcolor{blue}{Blue} and \textcolor{red}{red}  answers represent correct and incorrect predictions respectively.}
  \label{fig:case-study}
\end{figure*}
\subsection{Ablation Studies}
Besides the full model, we also ran several experiments on variants of LOGOS to examine the effectiveness of each component. Results are shown in Table \ref{tab:ablation}. We analyze model variant performance from three perspectives.

\paragraph{Effect of ROI Localization}
Rows \#1-\#3 show the effectiveness of ROI localization based on question information. We see an improvement when either the question-visual pretraining or the question-OCR modeling is involved, meaning that grounding questions in both modalities are helpful. Specifically, we see better performance ($\sim0.7\%$) for question-OCR modeling comparing to question-visual pretraining. This is expected because the question and OCR tokens share the same embedding space in the textual modality, which leads to more efficient relationship learning.

\paragraph{Effect of Using Multiple OCR Systems} 
Compared to simply selecting the highest-quality OCR system, LOGOS utilizes different capabilities of different OCR systems to contribute towards better performance. Rows \#3-\#5 show the results under the same modeling setting except for the source of OCR tokens. We can see that changing from a low-quality OCR system (Rosetta OCR) to a better one (Azure OCR) results in a huge gain ($\sim8.1\%$) in terms of accuracy, but using both and selecting with a selector results in further improvement ($\sim0.9\%$).

\begin{table}[h]
    \centering
    \scalebox{1.1}{
    \begin{tabular}{l|c}
    \toprule
        \textbf{Inference source} & \textbf{Accuracy} \\
    \midrule
        Rosetta Only & 43.26 \\
        Azure Only & 50.02 \\
        Select Better Confidence & 51.53 \\
    \midrule 
        \textbf{Selection Accuracy} & 71.96 \\
    \bottomrule
    \end{tabular}}
    \caption{Ablation study on the OCR Source Selection module. The values are obtained from the same model snapshot trained on TextVQA+STVQA.}
    \label{tab:selector}
\end{table}

In order to verify that the improvement is not only achieved by more diverse training data, we conduct an extra experiment by using both OCR sources during training but limiting the source during inference. The results are shown in Table \ref{tab:selector}. Among the whole validation dataset of TextVQA, there are 22.4\% of data where only one OCR system predicts the right answer, and in 71.96\% of these cases, LOGOS correctly selects the OCR source based on confidence.

\paragraph{Effect of Scene Text Clustering }
By comparing Row \#5 and \#6, we find that by grouping scene text within a similar area and adding the corresponding position representation, our model achieved an improvement of around 0.6\%. Note that all the methods mentioned in this paper are compatible with each other and can be used at the same time, by which we can obtain the state-of-the-art model with results reported in Row \#7 compared to models using only TextVQA as OCR data for training.

\subsection{Case Studies}
\label{sec:case-study}
In this section, we analyze predictions of different model variants of LOGOS on validation set questions to check module effectiveness. We show two examples in Figure~\ref{fig:case-study}. 

The case on the left demonstrates the effect of text clustering. The two images show OCR text pieces without and with text clustering respectively. Without the clustering information and with only the bounding box coordinates, the model still struggles to find the connection between individual words to form the phrase ``coloring with stain". Adding clustering-based word position information leads to much easier inference for the model to predict the correct answer. 

The case on the right further emphasizes the importance of utilizing detection results from multiple OCR sources. When one OCR source (Azure) incorrectly misses the key evidence text (the ``3" on the player's jersey), the model will most certainly generate an incorrect answer. After incorporating the source selection module, the model is able to dynamically choose the most reasonable source, based on its understanding of the question and relationship between text and objects. In this case, the model successfully grounds the question to the correct object (the jersey on the left) and answers the correct jersey number 3 instead of 17.


\begin{figure}[bht]
    \begin{center}
    \includegraphics[width=0.8\linewidth]{figure/err_case1.png} \\
    \end{center}
    {
    \textbf{Question:} How big is the TV? \\
    \textbf{LOGOS:} \textcolor{red}{8k} \\
    \textbf{Ground Truth:} \textcolor{blue}{98'' / 98 inch}} \\
    \begin{center}
        \includegraphics[width=0.9\linewidth]{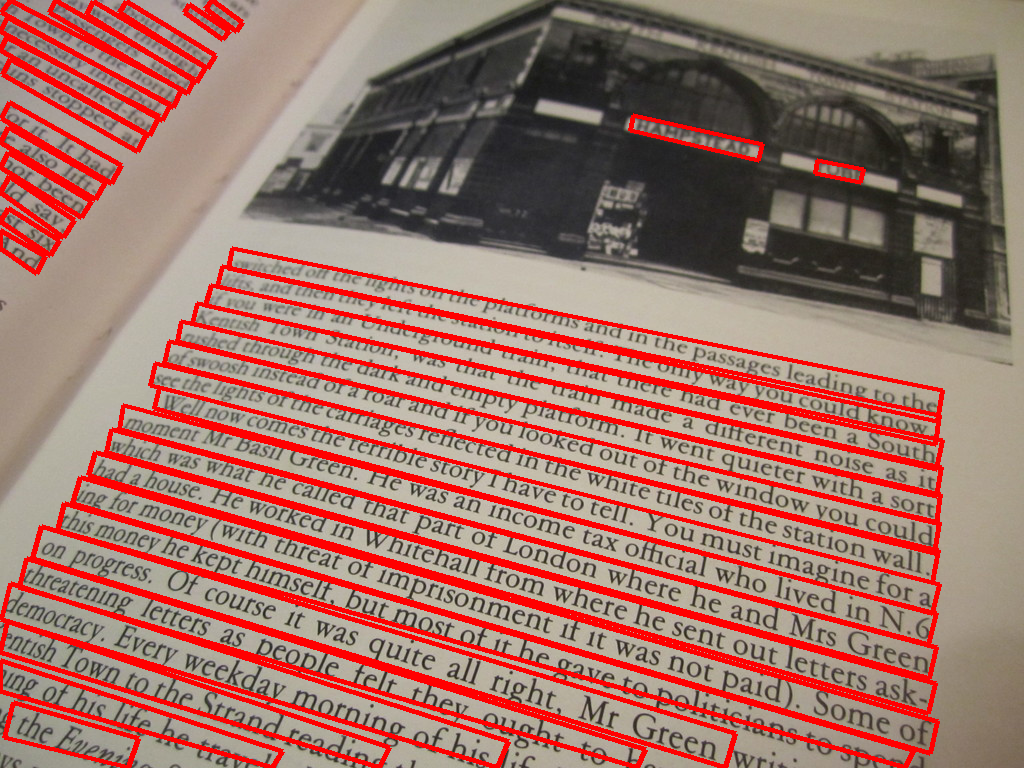} \\
    \end{center}
    {
    \textbf{Question:} What was Mr. Green's first name? \\
    \textbf{LOGOS:} \textcolor{red}{Hampstead} \\
    \textbf{Ground Truth:} \textcolor{blue}{Basil} 
    }
  \caption{Qualitative examples of LOGOS failed cases. \textcolor{blue}{Blue} and \textcolor{red}{red}  answers represent correct and incorrect predictions respectively.}
  \label{fig:err_case}
\end{figure}

\subsection{Error Analysis}
In this section, we present analysis of error types to see how our model achieved better accuracy and what limits further improvement. We randomly sampled 30 negative examples from M4C and another 30 from LOGOS and counted the number of errors caused by bad OCR quality, failure to group scene text, or failure to locate questions in the related area. The proportion of such errors reduced from 80.0\% to 53.3\%, which proves LOGOS's ability to better locate questions and utilize scene text clusters from multiple OCR systems.

When looking into negative examples from LOGOS, we also notice two typical types of error which we present in Figure~\ref{fig:err_case}. In these two cases, the model correctly recognizes and locates the question to the scene text, but still failed to generate the correct answer. 

In the first case, the question asks about the size of the TV. Answering such questions requires a QA model to be equipped with proper external knowledge. Here, the model needs to understand measurements and different kinds of notations such as \textit{8K} for resolution and \textit{98"} for size. Other types of external knowledge in similar TextVQA error cases include: recognizing time, identifying dates, and counting and calculation.

As for the second case, although all the words are detected correctly by the OCR system and grouped together, the model struggles to completely understand the long text.
Answering this type of question further requires the model to fully understand the content, which is almost impossible to be achieved without a more specific design, for example, a pipeline with a reading comprehension module.


\section{Conclusion and Future Work}
\label{sec:conclusion}

In this paper, we propose LOGOS, a novel model that hierarchically groups evidence from both image and text modality, and outputs answers from multiple noisy OCR systems. Results reveal that LOGOS outperforms the state-of-the-art models without using additional OCR training data. Detailed analysis shows that LOGOS can not only learn which region to focus on given the question but can also generate coherent answers with correct spatial order from the original image. 

Based on the analysis and comparison with other models, there are also some potential areas for future work. We do not use any external OCR-related datasets to enhance the model's ability to model scene text, and we expect a higher score when more OCR data are applied to our model. Our model makes the first attempt to better utilize scene text from images in the multi-modal setting. We observe a huge improvement on Text-VQA performance, and we believe this method can also be applied to similar tasks such as Text-Caption \cite{sidorov2020textcaps} where the understanding of scene text plays an important role.





{\small
\bibliographystyle{ieee_fullname}
\bibliography{custom}

\begin{thebibliography}{10}\itemsep=-1pt

\bibitem{almazan2014word}
Jon Almaz{\'a}n, Albert Gordo, Alicia Forn{\'e}s, and Ernest Valveny.
\newblock Word spotting and recognition with embedded attributes.
\newblock {\em IEEE transactions on pattern analysis and machine intelligence},
  36(12):2552--2566, 2014.

\bibitem{bigham2010vizwiz}
Jeffrey~P Bigham, Chandrika Jayant, Hanjie Ji, Greg Little, Andrew Miller,
  Robert~C Miller, Robin Miller, Aubrey Tatarowicz, Brandyn White, Samual
  White, et~al.
\newblock Vizwiz: nearly real-time answers to visual questions.
\newblock In {\em Proceedings of the 23nd annual ACM symposium on User
  interface software and technology}, pages 333--342, 2010.

\bibitem{biten2019scene}
Ali~Furkan Biten, Ruben Tito, Andres Mafla, Lluis Gomez, Mar{\c{c}}al Rusinol,
  Ernest Valveny, CV Jawahar, and Dimosthenis Karatzas.
\newblock Scene text visual question answering.
\newblock In {\em Proceedings of the IEEE/CVF International Conference on
  Computer Vision}, pages 4291--4301, 2019.

\bibitem{borisyuk2018rosetta}
Fedor Borisyuk, Albert Gordo, and Viswanath Sivakumar.
\newblock Rosetta: Large scale system for text detection and recognition in
  images.
\newblock In {\em Proceedings of the 24th ACM SIGKDD International Conference
  on Knowledge Discovery \& Data Mining}, pages 71--79, 2018.

\bibitem{deng2009imagenet}
Jia Deng, Wei Dong, Richard Socher, Li-Jia Li, Kai Li, and Li Fei-Fei.
\newblock Imagenet: A large-scale hierarchical image database.
\newblock In {\em 2009 IEEE conference on computer vision and pattern
  recognition}, pages 248--255. Ieee, 2009.

\bibitem{devlin2018bert}
Jacob Devlin, Ming-Wei Chang, Kenton Lee, and Kristina Toutanova.
\newblock Bert: Pre-training of deep bidirectional transformers for language
  understanding.
\newblock {\em arXiv preprint arXiv:1810.04805}, 2018.

\bibitem{gao2020structured}
Chenyu Gao, Qi Zhu, Peng Wang, Hui Li, Yuliang Liu, Anton van~den Hengel, and
  Qi Wu.
\newblock Structured multimodal attentions for textvqa.
\newblock {\em arXiv preprint arXiv:2006.00753}, 2020.

\bibitem{gao2020multi}
Difei Gao, Ke Li, Ruiping Wang, Shiguang Shan, and Xilin Chen.
\newblock Multi-modal graph neural network for joint reasoning on vision and
  scene text.
\newblock In {\em Proceedings of the IEEE/CVF Conference on Computer Vision and
  Pattern Recognition}, pages 12746--12756, 2020.

\bibitem{gurari2018vizwiz}
Danna Gurari, Qing Li, Abigale~J Stangl, Anhong Guo, Chi Lin, Kristen Grauman,
  Jiebo Luo, and Jeffrey~P Bigham.
\newblock Vizwiz grand challenge: Answering visual questions from blind people.
\newblock In {\em Proceedings of the IEEE Conference on Computer Vision and
  Pattern Recognition}, pages 3608--3617, 2018.

\bibitem{hahsler2019dbscan}
Michael Hahsler, Matthew Piekenbrock, and Derek Doran.
\newblock {dbscan}: Fast density-based clustering with {R}.
\newblock {\em Journal of Statistical Software}, 91(1):1--30, 2019.

\bibitem{han2020finding}
Wei Han, Hantao Huang, and Tao Han.
\newblock Finding the evidence: Localization-aware answer prediction for text
  visual question answering.
\newblock {\em arXiv preprint arXiv:2010.02582}, 2020.

\bibitem{hu2020iterative}
Ronghang Hu, Amanpreet Singh, Trevor Darrell, and Marcus Rohrbach.
\newblock Iterative answer prediction with pointer-augmented multimodal
  transformers for textvqa.
\newblock In {\em Proceedings of the IEEE/CVF Conference on Computer Vision and
  Pattern Recognition}, pages 9992--10002, 2020.

\bibitem{jiang2018pythia}
Yu Jiang, Vivek Natarajan, Xinlei Chen, Marcus Rohrbach, Dhruv Batra, and Devi
  Parikh.
\newblock Pythia v0. 1: the winning entry to the vqa challenge 2018.
\newblock {\em arXiv preprint arXiv:1807.09956}, 2018.

\bibitem{kafle2018dvqa}
Kushal Kafle, Brian Price, Scott Cohen, and Christopher Kanan.
\newblock Dvqa: Understanding data visualizations via question answering.
\newblock In {\em Proceedings of the IEEE conference on computer vision and
  pattern recognition}, pages 5648--5656, 2018.

\bibitem{kant2020spatially}
Yash Kant, Dhruv Batra, Peter Anderson, Alex Schwing, Devi Parikh, Jiasen Lu,
  and Harsh Agrawal.
\newblock Spatially aware multimodal transformers for textvqa.
\newblock {\em arXiv preprint arXiv:2007.12146}, 2020.

\bibitem{karatzas2015icdar}
Dimosthenis Karatzas, Lluis Gomez-Bigorda, Anguelos Nicolaou, Suman Ghosh,
  Andrew Bagdanov, Masakazu Iwamura, Jiri Matas, Lukas Neumann,
  Vijay~Ramaseshan Chandrasekhar, Shijian Lu, et~al.
\newblock Icdar 2015 competition on robust reading.
\newblock In {\em 2015 13th International Conference on Document Analysis and
  Recognition (ICDAR)}, pages 1156--1160. IEEE, 2015.

\bibitem{kembhavi2016diagram}
Aniruddha Kembhavi, Mike Salvato, Eric Kolve, Minjoon Seo, Hannaneh Hajishirzi,
  and Ali Farhadi.
\newblock A diagram is worth a dozen images.
\newblock In {\em European Conference on Computer Vision}, pages 235--251.
  Springer, 2016.

\bibitem{kembhavi2017you}
Aniruddha Kembhavi, Minjoon Seo, Dustin Schwenk, Jonghyun Choi, Ali Farhadi,
  and Hannaneh Hajishirzi.
\newblock Are you smarter than a sixth grader? textbook question answering for
  multimodal machine comprehension.
\newblock In {\em Proceedings of the IEEE Conference on Computer Vision and
  Pattern recognition}, pages 4999--5007, 2017.

\bibitem{krasin2017openimages}
Ivan Krasin, Tom Duerig, Neil Alldrin, Vittorio Ferrari, Sami Abu-El-Haija,
  Alina Kuznetsova, Hassan Rom, Jasper Uijlings, Stefan Popov, Andreas Veit,
  et~al.
\newblock Openimages: A public dataset for large-scale multi-label and
  multi-class image classification.
\newblock {\em Dataset available from https://github. com/openimages}, 2(3):18,
  2017.

\bibitem{krishna2016visual}
R Krishna et~al.
\newblock Visual genome: Connecting language and vision using crowdsourced
  dense image annotations. arxiv.
\newblock {\em arXiv preprint arXiv:1602.07332}, 2016.

\bibitem{li2020oscar}
Xiujun Li, Xi Yin, Chunyuan Li, Pengchuan Zhang, Xiaowei Hu, Lei Zhang, Lijuan
  Wang, Houdong Hu, Li Dong, Furu Wei, et~al.
\newblock Oscar: Object-semantics aligned pre-training for vision-language
  tasks.
\newblock In {\em European Conference on Computer Vision}, pages 121--137.
  Springer, 2020.

\bibitem{liu2020cascade}
Fen Liu, Guanghui Xu, Qi Wu, Qing Du, Wei Jia, and Mingkui Tan.
\newblock Cascade reasoning network for text-based visual question answering.
\newblock In {\em Proceedings of the 28th ACM International Conference on
  Multimedia}, pages 4060--4069, 2020.

\bibitem{mishra2013image}
Anand Mishra, Karteek Alahari, and CV Jawahar.
\newblock Image retrieval using textual cues.
\newblock In {\em Proceedings of the IEEE International Conference on Computer
  Vision}, pages 3040--3047, 2013.

\bibitem{mishra2019ocr}
Anand Mishra, Shashank Shekhar, Ajeet~Kumar Singh, and Anirban Chakraborty.
\newblock Ocr-vqa: Visual question answering by reading text in images.
\newblock In {\em 2019 International Conference on Document Analysis and
  Recognition (ICDAR)}, pages 947--952. IEEE, 2019.

\bibitem{rahman2020integrating}
Wasifur Rahman, Md~Kamrul Hasan, Sangwu Lee, Amir Zadeh, Chengfeng Mao,
  Louis-Philippe Morency, and Ehsan Hoque.
\newblock Integrating multimodal information in large pretrained transformers.
\newblock In {\em Proceedings of the conference. Association for Computational
  Linguistics. Meeting}, volume 2020, page 2359. NIH Public Access, 2020.

\bibitem{ren2015faster}
Shaoqing Ren, Kaiming He, Ross Girshick, and Jian Sun.
\newblock Faster r-cnn: Towards real-time object detection with region proposal
  networks.
\newblock {\em arXiv preprint arXiv:1506.01497}, 2015.

\bibitem{sidorov2020textcaps}
Oleksii Sidorov, Ronghang Hu, Marcus Rohrbach, and Amanpreet Singh.
\newblock Textcaps: a dataset for image captioning with reading comprehension.
\newblock In {\em European Conference on Computer Vision}, pages 742--758.
  Springer, 2020.

\bibitem{singh2019towards}
Amanpreet Singh, Vivek Natarajan, Meet Shah, Yu Jiang, Xinlei Chen, Dhruv
  Batra, Devi Parikh, and Marcus Rohrbach.
\newblock Towards vqa models that can read.
\newblock In {\em Proceedings of the IEEE/CVF Conference on Computer Vision and
  Pattern Recognition}, pages 8317--8326, 2019.

\bibitem{veit2016coco}
Andreas Veit, Tomas Matera, Lukas Neumann, Jiri Matas, and Serge Belongie.
\newblock Coco-text: Dataset and benchmark for text detection and recognition
  in natural images.
\newblock {\em arXiv preprint arXiv:1601.07140}, 2016.

\bibitem{vinyals2015pointer}
Oriol Vinyals, Meire Fortunato, and Navdeep Jaitly.
\newblock Pointer networks.
\newblock {\em arXiv preprint arXiv:1506.03134}, 2015.

\bibitem{wang2020general}
Xinyu Wang, Yuliang Liu, Chunhua Shen, Chun~Chet Ng, Canjie Luo, Lianwen Jin,
  Chee~Seng Chan, Anton van~den Hengel, and Liangwei Wang.
\newblock On the general value of evidence, and bilingual scene-text visual
  question answering.
\newblock In {\em Proceedings of the IEEE/CVF Conference on Computer Vision and
  Pattern Recognition}, pages 10126--10135, 2020.

\bibitem{wang2019words}
Yansen Wang, Ying Shen, Zhun Liu, Paul~Pu Liang, Amir Zadeh, and Louis-Philippe
  Morency.
\newblock Words can shift: Dynamically adjusting word representations using
  nonverbal behaviors.
\newblock In {\em Proceedings of the AAAI Conference on Artificial
  Intelligence}, volume~33, pages 7216--7223, 2019.

\bibitem{yang2020tap}
Zhengyuan Yang, Yijuan Lu, Jianfeng Wang, Xi Yin, Dinei Florencio, Lijuan Wang,
  Cha Zhang, Lei Zhang, and Jiebo Luo.
\newblock Tap: Text-aware pre-training for text-vqa and text-caption.
\newblock {\em arXiv preprint arXiv:2012.04638}, 2020.

\bibitem{zhu2020simple}
Qi Zhu, Chenyu Gao, Peng Wang, and Qi Wu.
\newblock Simple is not easy: A simple strong baseline for textvqa and
  textcaps.
\newblock {\em arXiv preprint arXiv:2012.05153}, 2020.

\end{thebibliography}
}

\appendix



\end{document}